\documentclass[10pt,twocolumn,letterpaper]{article}

\usepackage{3dv}
\usepackage{times}
\usepackage{epsfig}
\usepackage{graphicx}
\usepackage{amsmath}
\usepackage{amssymb}
\usepackage{amsthm}
\usepackage{booktabs}
\usepackage{bm}
\usepackage{caption}
\usepackage{color}
\usepackage{multirow}
\usepackage[pagebackref=true,breaklinks=true,letterpaper=true,colorlinks,bookmarks=false]{hyperref}

\newcommand{\myparagraph}[1]{\vspace{0.2em}\noindent\textbf{#1}}

\threedvfinalcopy 

\def\threedvPaperID{181} 

\ifthreedvfinal\fi
\definecolor{green}{rgb}{0.55, 0.71, 0.0}
\definecolor{amaranth}{rgb}{0.9, 0.17, 0.31}
\definecolor{amber}{rgb}{1.0, 0.49, 0.0}
\definecolor{azure}{rgb}{0.0, 0.5, 1.0}
\definecolor{byzantine}{rgb}{0.74, 0.2, 0.64}

\newcommand{\mysf}[1]{{\small \sf #1}}

\begin{document}
\title{PLACE: Proximity Learning of Articulation and Contact in 3D Environments}


\newcommand*{\affaddr}[1]{#1} 
\newcommand*{\affmark}[1][*]{\textsuperscript{#1}}
\newcommand*{\email}[1]{\small{\texttt{#1}}}

\author{
Siwei Zhang\affmark[1] \quad
Yan Zhang\affmark[1] \quad
Qianli Ma\affmark[1,2]  \quad
Michael J. Black\affmark[2]  \quad
Siyu Tang\affmark[1]\\
\affaddr{\affmark[1]ETH Z\"urich} \quad \affaddr{\affmark[2]Max Planck Institute for Intelligent Systems}\\
\email{\{siwei.zhang,yan.zhang,siyu.tang\}@inf.ethz.ch} \quad \email{\{qma,black\}@tue.mpg.de}
}


\twocolumn[{%
\renewcommand\twocolumn[1][]{#1}%
\maketitle
\begin{center}
    \newcommand{\teaserwidth}{\textwidth}
\vspace{-0.2in}
    \centerline{
    \includegraphics[width=1\linewidth]{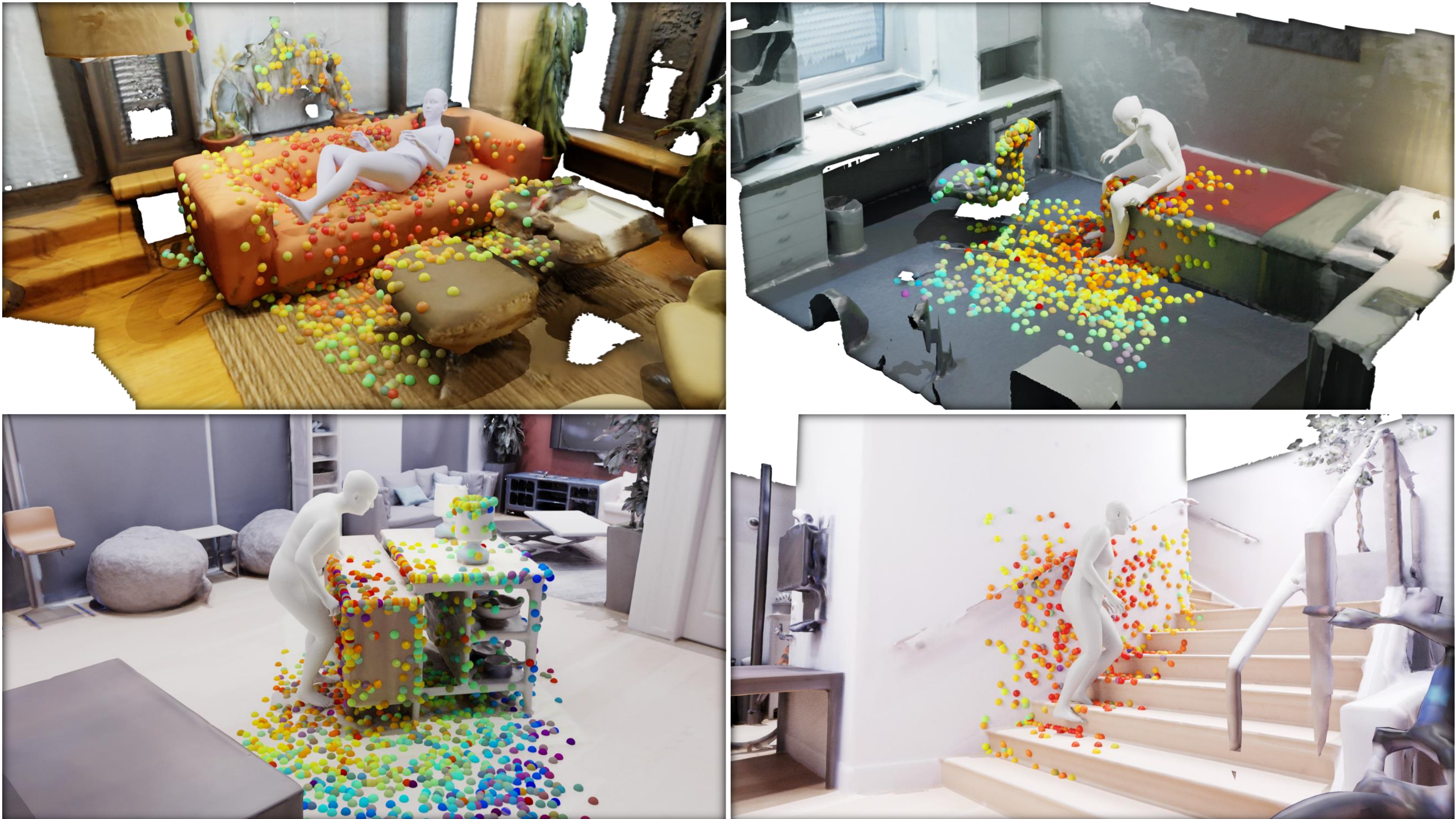}
     }
   \captionof{figure}{Given a 3D scene mesh, \textbf{PLACE} generates human-scene proximal relationships and infers a plausible human body. The selected local scene vertices are denoted by spheres, and the color denotes the generated minimum distance to the body surface. Distances are normalized to $[0,1]$ for visualization, and red to purple means close to far.}  
\label{fig:teaser}
\end{center}%
}]

\maketitle

\begin{abstract}
High fidelity digital 3D environments have been proposed in recent years, however, it remains extremely challenging to automatically equip such environment with realistic human bodies.
Existing work utilizes images, depth or semantic maps to represent the scene, and parametric human models to represent 3D bodies. 
While being straightforward, their generated human-scene interactions are often lack of naturalness and physical plausibility. 
Our key observation is that humans interact with the world through body-scene contact. 
To synthesize realistic human-scene interactions, it is essential to effectively represent the physical contact and proximity  between the body and the world. 
To that end, we propose a novel interaction generation method, named \emph{PLACE} (Proximity Learning of Articulation and Contact in 3D Environments), which explicitly models the proximity between the human body and the 3D scene around it. 
Specifically, given a set of basis points on a scene mesh, we leverage a conditional variational autoencoder to synthesize the minimum distances from the basis points to the human body surface. 
The generated proximal relationship exhibits which region of the scene is in contact with the person. 
Furthermore, based on such synthesized proximity, we are able to effectively obtain expressive 3D human bodies that interact with the 3D scene naturally. 
Our perceptual study shows that \emph{PLACE} significantly improves the state-of-the-art method, approaching the realism of real human-scene interaction.
We believe our method makes an important step towards the fully automatic synthesis of realistic 3D human bodies in 3D scenes. The code and model are available for research at \url{https://sanweiliti.github.io/PLACE/PLACE.html}.
\end{abstract}

\section{Introduction}
\label{sec:intro}

Automated synthesis of realistic humans posed naturally in a 3D scene is essential for many applications, such as augmented and virtual reality, special movie effect and synthetic dataset generation for machine learning algorithms.
To achieve high realism, manual animation of human-scene interactions is often required. 
Recent works \cite{li2019putting, zhang2020generating} have proposed methods for the fully automated generation of human bodies that interact with the 3D world, and yet the naturalness and the realism of their results are still far behind the captured real human-scene interaction such as \cite{PROX:2019}.

We argue that the primal reason is that current approaches lack an explicit interaction representation. Humans interact with the 3D scene through {\it contact}. 
To explicitly model the contact relationship between the body and the scene is key to the realism of the synthesized humans in the scene.
However, existing methods use images, semantic and depth maps to represent the 3D environment \cite{li2019putting, zhang2020generating}. While being easily integrated into deep convolutional networks, the 3D scene structure and the proximity between the body and the scene are not explicitly modeled, especially for the regions that are occluded from the camera view, making it hard to effectively enforce constraints in 3D, such as no inter-penetration and proper contact.

To address these problems, in this paper we propose explicit representations for modelling 3D scene geometry and person-scene contact relationships in a coherent manner. Inspired by the Basis Point Set (BPS) method \cite{Prokudin_2019_ICCV}, which efficiently represents a 3D human body by its minimum distances to a set of random points, we propose a {\em two-stage} BPS encoding scheme: given a 3D scene mesh and a fixed set of basis points in 3D space, we first compute the minimum distances from the basis points to the scene mesh vertices, and use this distance-based feature to represent the 3D scene. 
Furthermore, we use the selected scene vertices to compute their minimum distances to the human body surface, and use this distance-based feature to represent the person-scene interaction. Consequently, the scene representation and the person-scene interaction representation are modelled in a coherent manner regardless of different scene structures and body poses. 
Note that, our distance-based interaction representation explicitly captures the contact and the proximal relationships between the human body and the scene and can be regarded as the scene `affordance'. 
Compared with other person-scene interaction representations \cite{do2018affordancenet,qi2018human, roy2016multi,10.3389/fnbot.2020.00045, savva2016pigraphs}, ours is purely geometry-based, is more efficient than dense affordance maps, and requires neither scene semantics nor action type labeling, nor affordance class annotation.

With our novel person-scene representation and 3D scene encoding scheme, we propose a new generation method for placing 3D human bodies in 3D environments naturally. We name it as \textbf{PLACE} (Proximity Learning of Articulation and Contact in 3D Environments). Specifically, we learn a conditional variational autoencoder (cVAE) to generate plausible contact and proximal relationships between a human body and a 3D scene. 
We further infer a plausible full body mesh that complies with  the generated human-scene proximity.
Following \cite{PROX:2019, zhang2020generating}, we refine the body mesh via optimization; going beyond prior work, we introduce a novel objective term based on our interaction representation that further improve the performance.

We evaluate the performance of our method on three datasets: \textbf{PROX} \cite{PROX:2019}, \textbf{MP3D} \cite{Matterport3D,zhang2020generating}, and \textbf{Replica} \cite{replica19arxiv}, with diverse metrics, namely diversity, physical plausibility, and perceptual naturalness. Experiments show that our method outperforms state-of-the-art methods \cite{li2019putting, zhang2020generating}. Particularly, the extensive perceptual studies demonstrate that our method significantly improves the naturalness and the realism of the synthesized human-scene interactions, making an important step towards the fully automatic generation of realistic human bodies in 3D environments.

In summary, our contributions are:
(1) We propose a novel human-scene interaction representation that is efficient, compact and consistent across various bodies and scene structures. 
(2) Based on the proposed representation, we propose a generative model (\textbf{PLACE}) that is able to synthesize full human body meshes with natural poses and plausible contact relationships in novel scenes.
(3) We show through comprehensive experiments that \textbf{PLACE} achieves new state-of-the-art performance, without using additional modalities (e.g.~depth and mesh semantics) that are required in prior works \cite{li2019putting, zhang2020generating}.

\section{Related Work}
\label{sec:related_work}

\myparagraph{2D human-object interaction.} 
Perceiving human-object interactions in 2D images has been studied extensively \cite{delaitre2011learning, fang2018pairwise, gkioxari2018detecting, kato2018compositional, kim2020detecting, li2020detailed, li2019transferable,liu2019forecasting, shen2018scaling, wang2020learning,yao2010modeling,zhou2020cascaded}. 
Gkioxari et al.~\cite{gkioxari2018detecting} detect (human, verb, object) triplets using human appearance as cues to localize interacted objects. 
Fang et al.~\cite{fang2018pairwise} learn a pairwise body-part
attention model, which focuses on crucial body parts and their corresponding interactions. 
Wang et al.~\cite{wang2020learning} predict interaction points to localize and classify the interaction directly.
Li et al.~\cite{li2020detailed} learn a joint 2D-3D representation by estimating 3D human pose and interacted object position.
Besides detecting interactions, generating objects with natural interactions in 2D images is also studied. 
Tan et al.~\cite{tan2018and} learn a human instance composition model, which selects and transforms a human to an input image background. 
Lee et al.~\cite{lee2018context} insert an object or human instance mask into the semantic map of an image, and predicts the shape and pose of the object mask. 
Dvornik et al.~\cite{dvornik2019importance} predict which object is suitable to be placed in a given region in the image as a data augmentation method.

\myparagraph{3D human-scene interaction.}
The human-scene interaction in 3D has also been widely studied in the literature, e.g.~in \cite{caoHMP2020,grabner2011makes,gupta20113d, kim2014shape2pose,savva2014scenegrok, savva2016pigraphs,  zhu2016inferring}. 
Savva et al.~\cite{savva2016pigraphs} learn a joint distribution of human poses and object structures, and generate plausible human body poses and object arrangements given the action. 
Gupta et al.~\cite{gupta20113d} propose a human-centric scene representation to model physical interactions between the body and the given indoor scene, to predict human poses supported by the scene geometry. 
Kim et al.~\cite{kim2014shape2pose} predict a human pose on a given 3D object, by learning an affordance model for a class of shapes. 
Grabner et al.~\cite{grabner2011makes} learn an affordance detector to identify regions in a 3D scene that support a particular functional category such as `sitting'. 
Zhu et al.~\cite{zhu2016inferring} infer forces or pressures applied on the body parts interacting with the scene, and learn the comfort intervals by physical simulation. 
Monszpart et al.~\cite{monszpart2019imapper} recover both a plausible scene arrangement and human motions to fit an input monocular video by jointly reasoning about scene objects and human motions over space-time. 
Chen et al.~\cite{chen2019holistic++} jointly learn scene parsing, object bounding-boxes, camera pose, room layout and 3D human pose estimation given a single-view image. 
Li et al.~\cite{li2019putting} learn a 3D pose generative model to automatically put 3D body skeletons into the input scene represented by RGB, RGB-D, or depth image.
Cao et al.~\cite{caoHMP2020} predict 3D human paths and pose sequences given a scene image and 2D pose histories.
Most similar with our work, recently Zhang et al. \cite{zhang2020generating} propose a cVAE to generate 3D human mesh modeled by SMPL-X \cite{SMPL-X:2019} model given a 3D environment represented by RGB-D image. Taheri et al. \cite{taheri2020grab} use Basis Point Sets \cite{Prokudin_2019_ICCV} to encode the object and infer hand pose.

\myparagraph{Ours versus related work.}
Compared to previous work, we explicitly use the proximity to model human-scene interaction in 3D. With this compact representation, we train a cVAE from real human-scene interaction data to
synthesize the proximal relations between the human body and the scene. We show significant improvement on a large-scale perceptual study. The results indicate that our method is approaching the realism of real human-scene interaction.


\begin{figure*}[t!]
    \centering
    \includegraphics[width=\linewidth]{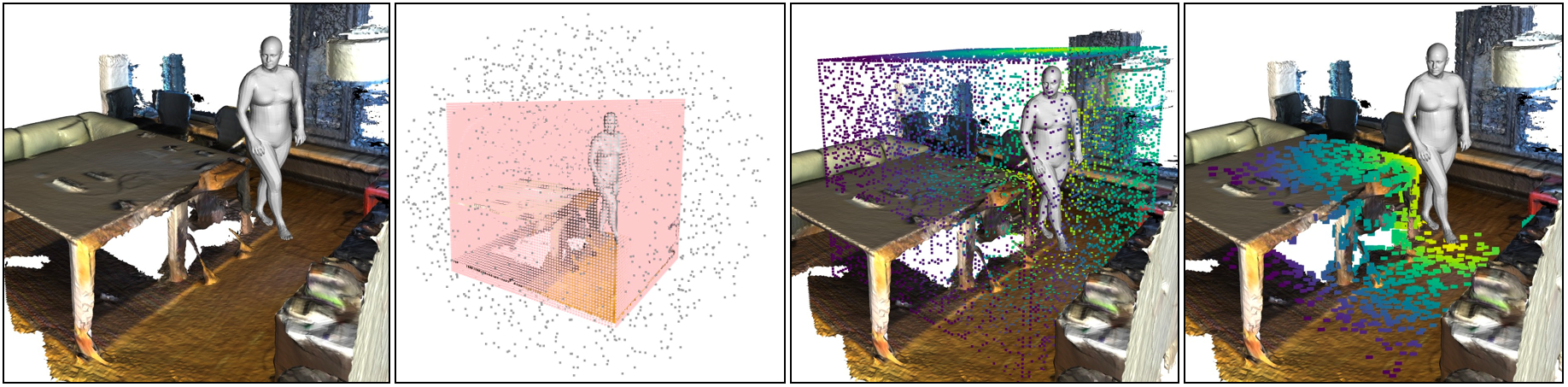}
    \caption{Illustration of human-scene interaction feature extraction (Sec.~\ref{sec:bps}). From left to right: (1) A pair of human-scene meshes from the \textbf{PROX} dataset \cite{PROX:2019}. (2) The cropped scene mesh as well as the 3D cage (in pink) are normalized into the unit sphere. (3) The derived scene BPS $\mathcal{V}_s$ is denoted by colored points, and the color indicates the distance-based body BPS features, i.e. the minimum distance from each scene basis point to the body mesh vertices. Small to large distance is denoted by bright yellow to dark purple. (4) The scene BPS and their minimum distances to the body only on the original scene mesh.}
    \label{fig:bps_feature_1}
\end{figure*}
\section{Proximity Representation for Interaction}
\label{sec:bps}
In this section, we introduce our novel proximity representation for modeling human-scene interaction.
Specifically, we use the Basis Point Sets (BPS) encoding scheme \cite{Prokudin_2019_ICCV} to represent the scene, the human body, and their interactions. The BPS method \cite{Prokudin_2019_ICCV} encodes a 3D object into a feature vector, which is defined on a fixed set of randomly selected basis points. This feature vector, i.e.~the distances from the basis points to their nearest neighbors on the target object, essentially captures the proximity relationship between the basis points and the object surface. Such property is well in line with our objective: by selecting basis points on the scene mesh, this distance-based feature naturally represents the body-scene contact and proximity.
Furthermore, in order to synthesize novel human bodies in an arbitrary 3D environment, we need an automatic way to select and encode the basis points on a scene mesh.
To that end, we propose a \textit{two-stage} BPS scheme that encodes the scene and the body consecutively (see Fig.~\ref{fig:bps_feature_1} for detailed illustration).

\myparagraph{Stage 1: BPS encoding for 3D Scene.}
To represent a 3D scene, we follow \cite{Prokudin_2019_ICCV} and create a set of $N=10^4$ uniformly random basis points $\mathcal{V}_{o}\subset \mathbb{R}^{N\times3}$ within a unit sphere.  $\mathcal{V}_{o}$ are kept fixed throughout both the training and testing pipeline for all 3D scenes. The scene meshes are also normalized into the unit sphere.
Then, for each basis point in $\mathcal{V}_{o}$, we find its nearest neighbor in the normalized scene mesh vertices, select the corresponding scene vertex and compute the distance.
Such minimum distances form the \textit{scene feature}  ${\bm x}_{s}\in\mathbb{R}^{N\times 1}$; and the selected $N$ scene vertices, $\mathcal{V}_{s}\subset\mathbb{R}^{N\times3}$, will serve as \textit{scene BPS} to encode the body in the second stage.

In practice, 
only the part of the scene that is close to the human body is most relevant for modeling the interaction.
This motivates us to consider `local scenes', i.e.~encode only part of the scene surrounding the body. To do so, we specify a virtual 3D cage around the body, and only consider the scene vertices within the cage. Furthermore, we regularly sample points on the top and the 4 side faces of the cage, and consider them as the `ceiling' and `walls' of the local scene. The local scene then goes through the encoding pipeline described above.
We find the inclusion of the virtual ceiling and walls useful in many cases, as otherwise the scene BPS $\mathcal{V}_{s}$ can fail to capture the full human body. 
For example, in the simplest configuration where the human stands in a 3D environment that only consists a floor, without the virtual ceiling and the walls, $\mathcal{V}_{s}$ only consists the vertices on the floor, then the body except the feet will hardly be encoded by the scene BPS.

\myparagraph{Stage 2: BPS encoding for 3D Body.}
The obtained scene BPS $\mathcal{V}_{s}$, which also includes vertices on the cage surfaces, is employed to encode the body mesh, following the same nearest neighbor encoding pipeline. 
We use the SMPL-X \cite{SMPL-X:2019} model to represent the 3D human body as mesh with vertices  $\mathcal{V}_b\subset\mathbb{R}^{10475\times3}$. The BPS encoding compresses $\mathcal{V}_{b}$ into the \textit{body feature}, ${\bm x}_{b}\in\mathbb{R}^{N\times 1}$, which is obtained by computing the distance between each point in scene BPS $\mathcal{V}_{s}$ to its nearest point of the body mesh vertices. The proposed \textit{body feature} is scene dependent, and explicitly encodes the contact and the proximal relations between the human body and the scene. Therefore it can be used to as the interaction representation to model and synthesize realistic human bodies in a 3D scene. 
Note that the basis points in the unit sphere (used in \textbf{Stage 1}) are always fixed. Consequently, 

every dimension in the scene feature always corresponds to the same dimension in the body feature. Therefore the proposed scene-body representation is consistent across various human body poses and scene structures.

\section{PLACE}
Given the proposed proximity representation for a pair of interacting human body and 3D scene, here we introduce our generative model, PLACE, which takes the scene BPS $\mathcal{V}_s$, the scene feature ${\bm x}_s$, and the body feature ${\bm x}_b$ as input to synthesize plausible human-scene contact and proximal relationships and to infer expressive 3D human bodies that interact with the 3D scene naturally.

\begin{figure}[t!]
    \centering
    \includegraphics[width=\linewidth]{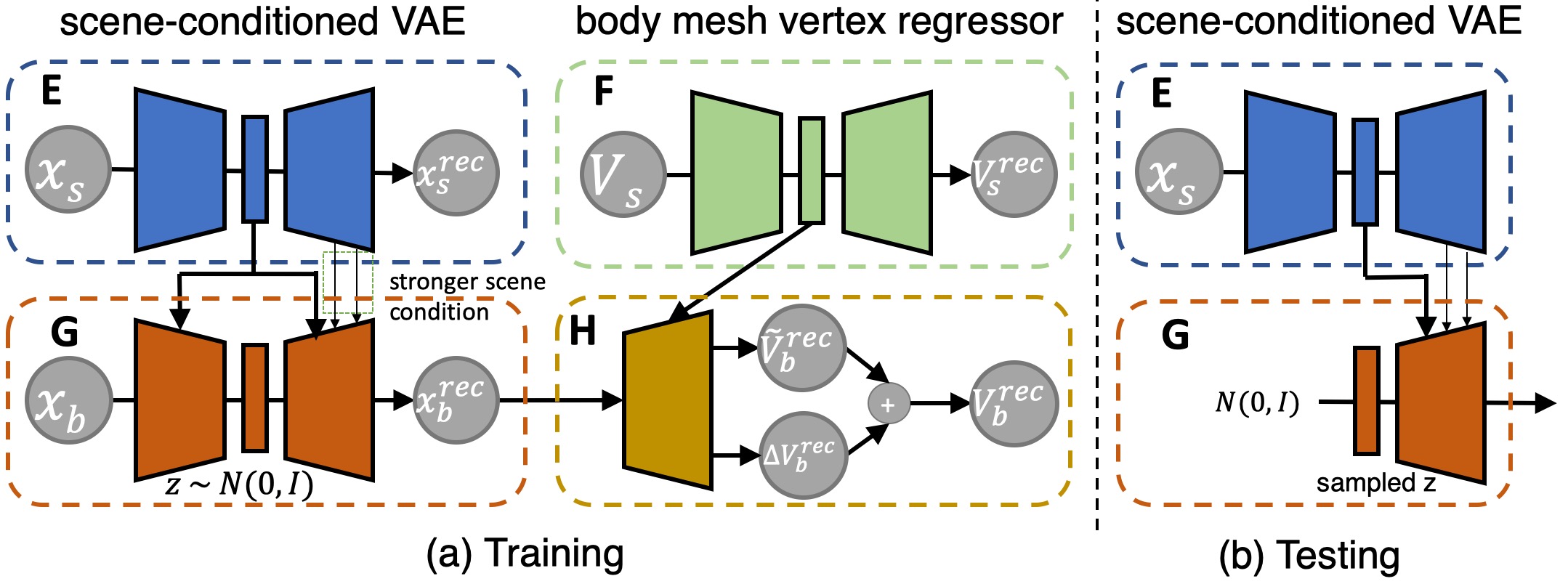}
    \caption{Illustration of the network architecture for PLACE. The training pipeline is illustrated in (a). ${\bm x}_s$, $\mathcal{V}_s$, ${\bm x}_b$ and $\mathcal{V}_b$ denote the scene feature, the scene BPS, the body feature, and the full body mesh vertices, respectively. ${\Delta \mathcal{V}}^{rec}_b$ denotes a global 3D translation for all the body vertices. 
    Note that all point coordinates are relative to the center of the normalized 3D cage in the unit sphere. The ``stronger scene condition'' introduces additional intermediate features to the decoder of $G$. During inference stage, the scene-conditioned VAE is shown in (b), with the same body regressor as in training stage.}
    \label{fig:network}
\end{figure}

\myparagraph{Network Architecture}
The network architecture is illustrated in Fig.~\ref{fig:network}. It incorporates a conditional variational autoencoder (cVAE) \cite{kingma2013auto, sohn2015learning} to generate body features based on a scene feature auto-encoder $E$. The generator $G$ is followed by a multilayer perceptron $H$ as the regressor to produce full body meshes, based on the reconstructed body features and the scene BPS auto-encoder $F$. 
Intermediate features from the decoder of $E$ are introduced to the decoder of $G$ for better scene conditioning. 
Details of the network architecture and its ablation study are demonstrated in the Appendix (Fig.~\ref{fig:supp-network-detail} and Tab.~\ref{tab:supp-ablation}). 

\subsection{Losses and Learning}
Following Sec.~\ref{sec:bps}, we obtain a set of representations from each ground truth sample, namely $\{{\bm x}_{s},{\bm x}_{b},\mathcal{V}_{s},\mathcal{V}_{b}\}$ denoting the scene feature, the body feature, the scene BPS, and the full body vertices, respectively.
Note that the point coordinates are rescaled in the unit sphere, and set relative to the rescaled 3D cage center. We denote the operations of transforming from the rescaled 3D cage back to the original world as $\pi(\cdot)$, and its inverse operation as $\pi^{-1}(\cdot)$.
The training loss is given by
\begin{equation}
\label{eqn:training loss}
    \mathcal{L} = \mathcal{L}_{rec} + \alpha_1 \mathcal{L}_{KL} + \alpha_2 \mathcal{L}_{coll} + \alpha_3 \mathcal{L}_{contact},
\end{equation}
in which $\{\alpha_1, \alpha_2, \alpha_3\}$ are loss weights as hyper-parameters.

\myparagraph{Reconstruction term:} This term is given by
\begin{equation}
    \mathcal{L}_{rec} = |{\bm x}_{s} - {\bm x}^{rec}_{s}| + |{\bm x}_{b} - {\bm x}^{rec}_{b}| + |\mathcal{V}_b - \Tilde{\mathcal{V}}_b^{rec}|,
\end{equation}
in which the last term denotes vertex-wise reconstruction.

\myparagraph{KL-divergence term:} Following \cite{kingma2013auto}, we specify the latent prior as the standard normal distribution, and employ the same re-parameterization approach to model the inference posterior $q({\bm z}|{\bm x})$. To effectively avoid posterior collapse, we use the robust KL-divergence term as in \cite{zhang2020perpetual}. Our KL-divergence term is given by
\begin{equation}
    \mathcal{L}_{KL} = \Psi\left(q({\bm z}|{\bm x}) || \mathcal{N}(0, {\bm I}) \right),
\end{equation}
in which $\Psi(\cdot)$ is the Charbonnier function $\Psi(s) = \sqrt{s^2+1}-1$ \cite{charbonnier1994two}. According to \cite{zhang2020perpetual}, the smaller is the KL-divergence between the inference posterior and the prior, the smaller are the gradients to update this KL-divergence. 

\myparagraph{The collision term and the contact term:} 
As in \cite{zhang2020generating}, these two terms are given by
\begin{align}
\label{eq:loss_hs}
    \mathcal{L}_{coll} &= \frac{1}{|\mathcal{V}_b|} \sum_{{\bm v}_b \in \pi(\mathcal{V}^{rec}_b) }  |\Phi^{-}({\bm v}_b )|, \text{ and}\\
    \mathcal{L}_{contact} &=  \sum_{{\bm v}_b \in C(\pi(\mathcal{V}^{rec}_b)) } \min_{{\bm v}_s \in \mathcal{M}_{s}} \rho( |{\bm v}_b - {\bm v}_s| ),
\end{align}
in which $C(\cdot)$ denotes the mask to select body part for contact, $|\mathcal{V}_b|$ denotes the constant number of body mesh vertices, $\Phi^{-}(\cdot)$ is the scene negative signed distance field (SDF), $\mathcal{M}_{s}$ is the provided scene mesh, and $\rho(\cdot)$ is the Geman-McClure error function to remove the influence of scene mesh vertices far away from the body \cite{PROX:2019, zhang2020generating}. In contrast to \cite{PROX:2019, zhang2020generating}, the contact mask $C(\cdot)$ only includes feet, which we find is better to avoid body floating.

\subsection{Placing People in Novel Scenes}
Given a 3D scene without humans, the generation procedures are: (1) We randomly create a 3D cage within the scene and transform it into the pre-defined unit sphere. (2) Within the unit sphere, we compute the scene BPS $\mathcal{V}_s$ and the scene feature ${\bm x}_s$. (3) Given $z$ sampled from $\mathcal{N}(0, I)$, the cVAE is employed to generate a body feature ${\bm x}_b$ (as illustrated in Fig.~\ref{fig:network} (b)). (4) Based on $\mathcal{V}_s$ and ${\bm x}_b$, the regressor is employed to produce a set of body mesh vertices, which is then transformed back to the original world.

\myparagraph{Interaction-based Optimization.}
The body vertices produced by the regressor $H$ are not guaranteed to form a valid body shape: The raw output from the network can suffer from noisy surface geometry or inter-penetration with the scene. Therefore, we conduct an additional optimization step to convert the regressed body mesh vertices to a SMPL-X body mesh. Given the scene mesh $\mathcal{M}_s$, the body feature ${\bm x_b}$, and the regressed body mesh vertices $\mathcal{V}_b$, the optimization loss is given by
\begin{equation}
\label{eqn:optimization loss}
    \begin{split}
        \mathcal{L} (\theta) &= |\pi(\mathcal{V}_b) - \mathcal{V}_\textrm{SMPLX}(\theta)| + |{\bm x_b} - f(\pi^{-1} (\mathcal{V}_\textrm{SMPLX}(\theta))| \\
        &+ \lambda_1 \mathcal{L}_{coll} + \lambda_2 \mathcal{L}_{contact}
        + \lambda_3 \mathcal{L}_{vposer} \\
        &+ \lambda_4 \mathcal{L}_{hand} + \lambda_5 \mathcal{L}_{shape}, 
    \end{split}
\end{equation}
in which $\mathcal{V}_\textrm{SMPLX}(\theta)$ denotes vertices of the SMPL-X body mesh with the attribute $\theta$ (including global translation, orientation, body shape, pose and hand pose), $\mathcal{L}_{coll}$ and $\mathcal{L}_{contact}$ are the same in Eq.~\eqref{eq:loss_hs}. In addition, $\mathcal{L}_{vposer}$, $\mathcal{L}_{hand}$, $\mathcal{L}_{shape}$ are the VPoser loss \cite{PROX:2019, SMPL-X:2019}, L2 prior loss for hand and shape parameters in SMPL-X model to ensure natural body, hand poses and shapes of the optimized body, respectively. 
$\lambda$'s are loss weights as hyper-parameters, and $f(\cdot)$ is the operation to extract the body feature from the body mesh, demonstrated in Sec.~\ref{sec:bps}.

In the optimization loss, the first term denotes the vertex-wise reconstruction, in order to discourage dramatic deviation of the resultant SMPL-X body mesh from the regressed body vertices. Additionally, we employ the second term to encourage the resultant SMPL-X body mesh have the same contact relation with the scene. Unlike the heuristic contact term $\mathcal{L}_{contact}$, this BPS feature reconstruction term is data-driven, since the body feature can be learned from data and generated by our cVAE model. 

\begin{figure}[t!]
    \centering
    \includegraphics[width=\linewidth]{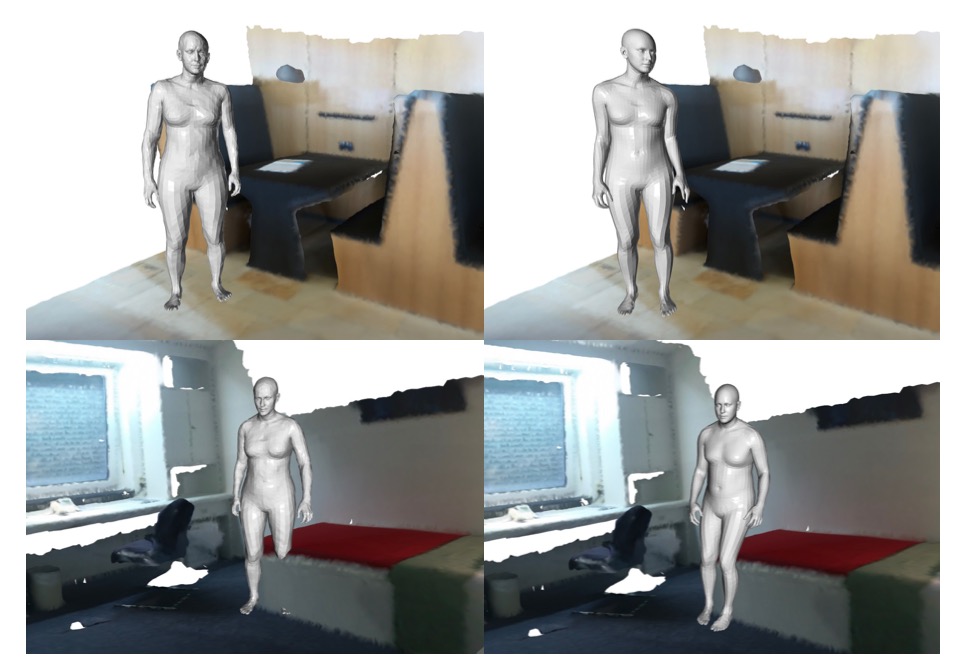}
    \caption{Illustration of optimization results. The left column shows two generated body mesh vertices, and the right column shows their optimized results.}
    \label{fig:opt_vis}
\end{figure}

Fig.~\ref{fig:opt_vis} shows two results of optimization. We can see that the human-scene contact is preserved, and their inter-penetration is considerably alleviated. 

\section{Experiments}
\label{sec:experiment}

\subsection{Datasets}
\myparagraph{PROX \cite{PROX:2019}} includes 12 different room 3D scans, and captures natural actions of 20 subjects, which are represented by SMPL-X body meshes \cite{SMPL-X:2019}. 
Following \cite{zhang2020generating}, we select `MPH1Library', `MPH16', `N0SittingBooth' and `N3OpenArea' for testing, and use the rests for training, from which we extract interaction features from 63,599 human-scene pairs.
To improve the generalization ability to new scenes, we perform data augmentation 4 times for each human-scene pair, and obtain 254,396 training samples in total.
During testing, we randomly set 1200 3D cages in each test scene, and generate one random sample in each cage. This results in 4800 examples for evaluation.

\myparagraph{MP3D \cite{Matterport3D}} has 90 building-scale scene scans with semantic annotations. Following the setup in \cite{zhang2020generating}, we perform testing on the same 7 scenes with corresponding scene SDFs, and generate the same number of random samples for each scene as in \cite{zhang2020generating}, obtaining 6400 random samples in total.

\myparagraph{Replica \cite{replica19arxiv}} contains reconstructed indoor scenes of clean dense geometry.
It includes 18 scenes of 5 different room types, among which we select 1 room in each type, resulting in 5 rooms (`office\_2', `hotel\_0', `room\_0', `frl\_apartment\_0', `apartment\_1') for testing. 1200 random samples are generated on each scene for evaluation.

\subsection{Our method and Baselines}
In this paper, we denote our model trained without $\mathcal{L}_\textrm{coll}$ and $\mathcal{L}_\textrm{contact}$ as \mysf{Ours}. 
Its ablation study is referred to Sec.~\ref{sec:ablation} and Appendix (Sec.~\ref{append_ablation_study}). We compare our method with two baselines:
(1) \mysf{PSI}~\cite{zhang2020generating}, which generates SMPL-X parameters given the scene depth and the semantic segmentation. Specifically, we perform comparison with their S1 model without the human-scene interaction training loss, both before and after the geometry-aware fitting, which are reported to outperform other methods. 
(2) \mysf{Li et al.}~\cite{li2019putting}, which generates 3D body stick figures based on scene appearance and depth. We perform comparison with its modified version proposed in \cite{zhang2020generating}, which generates SMPL-X human bodies for fair comparison.

It is noted that these two baselines generate SMPL-X parameters, yet our method directly produces body vertices, which might not form a valid body shape. To fairly compare with the baselines without the geometry-aware fitting step, we perform an optimization step to derive SMPL-X parameters from the body vertices, for which $\mathcal{L}_\textrm{contact}$ and $\mathcal{L}_\textrm{coll}$ in Eq.~\ref{eqn:optimization loss} are excluded. We denote this optimization as \mysf{SimOptim}, and the optimization with all terms as \mysf{AdvOptim} in the following content. Moreover, we denote the scene geometry-aware fitting proposed in \cite{zhang2020generating} as $\mathcal{L}_f$.

\subsection{Quantitative Evaluation}

We evaluate the performance in terms of perceptual naturalness, diversity, and physical plausibility, following Zhang et al. \cite{zhang2020generating}.

\subsubsection{Perceptual Naturalness}
We run perceptual studies on Amazon Mechanical Turk (AMT), and use two protocols to perform the comparison. For each scene and each model, we generate 100 and 400 examples for \textbf{PROX} and \textbf{MP3D} respectively, as in \cite{zhang2020generating}. 

\myparagraph{Unary user study. }
Like \cite{zhang2020generating}, we ask turkers to evaluate each individual generated result or ground truth, giving a score from 1 (strongly not natural) to 5 (strongly natural). Each example is rated by 5 turkers. 
The results are shown in Tab.~\ref{tab:user-study}. On both \textbf{PROX} and \textbf{MP3D}, our methods with \mysf{SimOptim} and \mysf{AdvOptim} outperform the baseline methods after their geometry-aware fitting. 

\myparagraph{Binary user study.}
To perform a direct comparison, we follow \cite{li2019putting,ma2020learning}, and show results of two methods to the turkers at the same time. The turkers are asked to pick the one that they think is more perceptually natural. 
For \textbf{PROX}, we compare the best instance of our method (the one trained with $\mathcal{L}_\textrm{contact}$ and \mysf{AdvOptim}) against \cite{zhang2020generating} with the geometry-aware fitting $\mathcal{L}_f$, and against the ground truth, respectively. For \textbf{MP3D}, we compare with \cite{zhang2020generating} with $\mathcal{L}_f$.

As shown in Tab.~\ref{tab:user-study-compare}, Our method outperforms \mysf{PSI} \cite{zhang2020generating} on both datasets by a large margin. More interestingly, in the direct comparison with the \textbf{PROX} ground truth, our generated results are regarded as more realistic by nearly half ($48.5\%$) of the users, indicating that our results are hardly distinguishable from the real human-scene interaction. It suggests that our method makes an important step towards the fully automatic synthesis of realistic 3D human bodies in 3D scenes.

\begin{table}[t]
\centering
\caption{\textbf{Unary user study.} We calculate the results from all turkers, and report the the scores w.r.t. the average $\pm$ the standard deviation. Best results except the ground truth are shown in boldface.}
\begin{tabular}{lcc}
  \toprule
  model &  \textbf{PROX} & \textbf{MP3D}   \\
  \midrule
  \mysf{Li et al.} \cite{li2019putting} + $\mathcal{L}_f$ & 3.37 $\pm$ 1.31 & 3.52 $\pm$ 1.34  \\
  \mysf{PSI} \cite{zhang2020generating} + $\mathcal{L}_f$ & 3.47 $\pm$ 1.27 & 3.23 $\pm$ 1.40  \\
  \mysf{Ours} + $\mathcal{L}_\textrm{contact}$ + \mysf{SimOptim} & 3.61 $\pm$ \textbf{1.16} & 3.77 $\pm$ 1.15 \\
  \mysf{Ours} + $\mathcal{L}_\textrm{contact}$ + \mysf{AdvOptim} &  \textbf{3.71} $\pm$ 1.17 & \textbf{3.84} $\pm$ \textbf{1.11} \\
  \midrule
  Ground truth& 3.94 $\pm$ 0.99 & -- \\
  \bottomrule
 
 \end{tabular}
\label{tab:user-study}
\end{table}    

\begin{table}[t]
\centering
\caption{\textbf{Binary user study.} Numbers show the percentage of the users that rate the corresponding method as more realistic.}
\begin{tabular}{lcc}
  \toprule
     Datasets & \multicolumn{2}{c}{\% users rated as ``better''} \\
    \midrule
        \multirow{2}{*}{\textbf{MP3D}} & \mysf{Ours} + $\mathcal{L}_\textrm{contact}$ + \mysf{AdvOptim} & \mysf{PSI} + $\mathcal{L}_f$ \\
                           & \textbf{70.1\%} & 29.9\%  \\
    \midrule
    \multirow{2}{*}{\textbf{PROX}} & \mysf{Ours} + $\mathcal{L}_\textrm{contact}$ + \mysf{AdvOptim} & \mysf{PSI} + $\mathcal{L}_f$ \\
                           & \textbf{69.0\%} & 31.0\% \\
    \midrule
    \multirow{2}{*}{\textbf{PROX}} & \mysf{Ours} + $\mathcal{L}_\textrm{contact}$ + \mysf{AdvOptim} & GT \\
                           & 48.5\% & \textbf{51.5\%} \\
  \bottomrule
 
 \end{tabular}
\label{tab:user-study-compare}
\end{table}

\subsubsection{Diversity}
\myparagraph{Evaluation metrics.} Like \cite{zhang2020generating}, we perform K-means to cluster the SMPL-X parameters of generated bodies into 20 clusters, and evaluate the model by: 1) the entropy of the cluster ID histogram for all samples, and 2) the cluster size, which denotes the average distance between the sample and the corresponding cluster center. The higher the better for both metrics. We report average scores over all test samples.

\myparagraph{Results.} 
Tab.~\ref{tab:diversity} presents the diversity results. Comparing with baselines, \mysf{Ours}+$\mathcal{L}_\textrm{contact}$+\mysf{SimOptim} achieve comparable entropy and cluster size on both \textbf{PROX} and \textbf{MP3D}. When applying \mysf{AdvOptim}, the cluster size is further increased by a large margin, indicating that the generated bodies are more scattered in the scene. Since the cluster size is correlated with the test scene size, its values are higher on \textbf{MP3D} and \textbf{Replica} than \textbf{PROX} due to larger room sizes.

\begin{table*}[h!]
\centering
\footnotesize
\caption{\textbf{Evaluation on diversity and physical plausibility}: For the entropy, the cluster size, the non-collision score and the contact score, their values are the higher the better. Row-wisely, the first part and the second part shows the comparison to baseline methods before and after geometry-aware fitting, respectively. For each part, the best results are in boldface.}
\begin{tabular}{lcccccccccccc}
  \toprule[1pt]
  & \multicolumn{6}{c}{Diversity} & \multicolumn{6}{c}{Physical plausibility} \\
  \cmidrule(lr){2-7} \cmidrule(lr){8-13}
  & \multicolumn{3}{c}{entropy} & \multicolumn{3}{c}{cluster size} & \multicolumn{3}{c}{non-coll} & \multicolumn{3}{c}{contact} \\
  \cmidrule(lr){2-4} \cmidrule(lr){5-7} \cmidrule(lr){8-10} \cmidrule(lr){11-13} 
  Methods &  \textbf{PROX} & \textbf{MP3D} & \textbf{Replica} & \textbf{PROX} & \textbf{MP3D} & \textbf{Replica} &  \textbf{PROX} & \textbf{MP3D} & \textbf{Replica} &  \textbf{PROX} & \textbf{MP3D} & \textbf{Replica} \\
  \midrule

  \mysf{Li et al.}~\cite{li2019putting} & 2.89 & 2.93 & - & 1.49  & 1.84 & - & 0.89 & 0.92 & - & 0.93 & 0.78 & - \\ 
  \mysf{PSI}~\cite{zhang2020generating}  & \textbf{2.96} & \textbf{2.99} & - & 2.51 & 2.81 & - & 0.93 & 0.94 & - & \textbf{0.95} & \textbf{0.80} & - \\
  \mysf{Ours} + $\mathcal{L}_\textrm{contact}$ & -& -& -& -& -& - & 0.95 & \textbf{0.97} & \textbf{0.92} & 0.90 & 0.57 & 0.90 \\
  \begin{tabular}{@{}@{}c}  \mysf{Ours} + $\mathcal{L}_\textrm{contact}$ \\~~+~\mysf{SimOptim} \end{tabular}  & 2.91 & 2.92 & 2.93 & \textbf{2.59} & \textbf{2.93} & 2.99 & \textbf{0.96} & \textbf{0.97} & 0.91 & \textbf{0.95} & 0.62 & \textbf{0.92} \\
  
  \midrule

  \mysf{Li et al.} + $\mathcal{L}_f$ & 2.93 & 2.92 & - & 1.52 & 1.94 & -  & 0.93 & 0.97 & - & \textbf{0.99} & \textbf{0.89} & - \\  
  \mysf{PSI} + $\mathcal{L}_f$ & \textbf{2.97} & \textbf{2.98} & - & 2.53 & 2.86 & - & 0.94 & 0.97 & - & \textbf{0.99} & 0.88 & - \\
  \begin{tabular}{@{}@{}c} \mysf{Ours} + $\mathcal{L}_\textrm{contact}$ \\~~+~\mysf{AdvOptim} \end{tabular} & 2.91 & 2.93 & 2.92 & \textbf{2.72} & \textbf{3.07} & 3.12 & \textbf{0.98} & \textbf{0.98} & 0.93 & \textbf{0.99} & 0.57 & 1.00 \\
  \bottomrule[1pt]
 \end{tabular}
\label{tab:diversity}
\end{table*}

\subsubsection{Physical Plausibility}
\myparagraph{Evaluation metrics.}
We evaluate the non-collision and contact scores between the generated body and scene mesh, following the metrics in \cite{zhang2020generating}.
Specifically, the non-collision score is the ratio of the number of body vertices with non-negative scene SDF values to the number of all body vertices (10475). The contact score is 1 if any body mesh vertex has a non-positive scene SDF value, and 0 otherwise. Then we report the average non-collision scores and average contact scores of all test examples.
Thus, a higher non-collision score indicates fewer body-scene inter-penetration, while a higher contact score suggests more contacts between the generated body and scene mesh. Higher values for both scores are desirable. 

\myparagraph{Results.}
Tab.~\ref{tab:diversity} shows the results on physical plausibility. On \textbf{PROX}, \mysf{Ours}+$\mathcal{L}_\textrm{contact}$+\mysf{SimOptim} achieves higher non-collision score and comparable contact score with the baselines, indicating that our proposed method can effectively alleviate scene-body inter-penetration, while preserving plausible contact relations.
When additionally employing the interaction-based optimization, performance is further consistently improved.
In addition, we find the SDFs on \textbf{MP3D} are not sufficiently reliable probably due to noisy scans\footnote{Illustrations are presented in the Appendix (Fig.~\ref{fig:supp-mp3d-sdf-example}).}. Thus, we further perform evaluation on the \textbf{Replica} dataset, from which high-quality scene SDFs can be derived. The results on \textbf{Replica} verify the effectiveness of our model, which is consistent with the performances on {\bf PROX}.

\subsection{Qualitative Results.}

\begin{figure}[t]
    \centering
    \includegraphics[width=\linewidth]{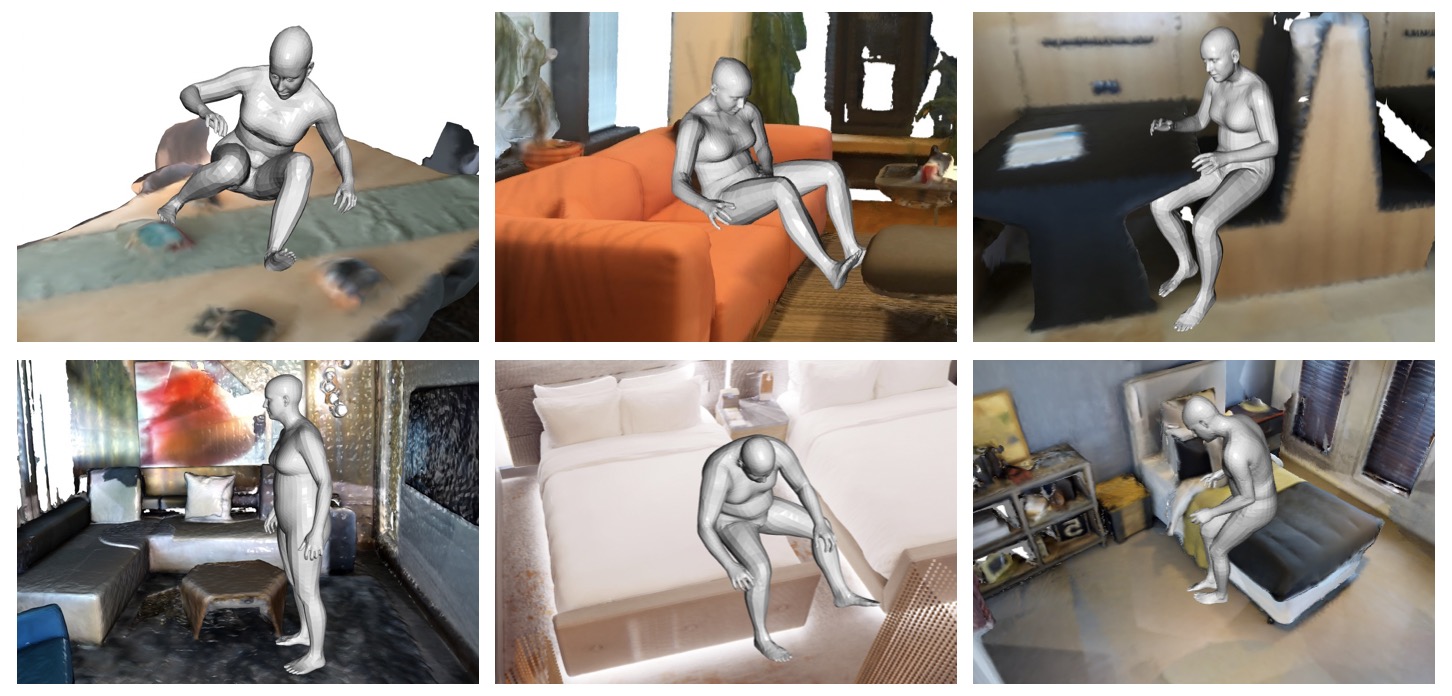}
    \caption{Generated results on \textbf{PROX}, \textbf{MP3D} and \textbf{Replica}.}
    \label{fig:generated_sample}
\end{figure}

\begin{figure*}[h!]
    \centering
    \includegraphics[width=\linewidth]{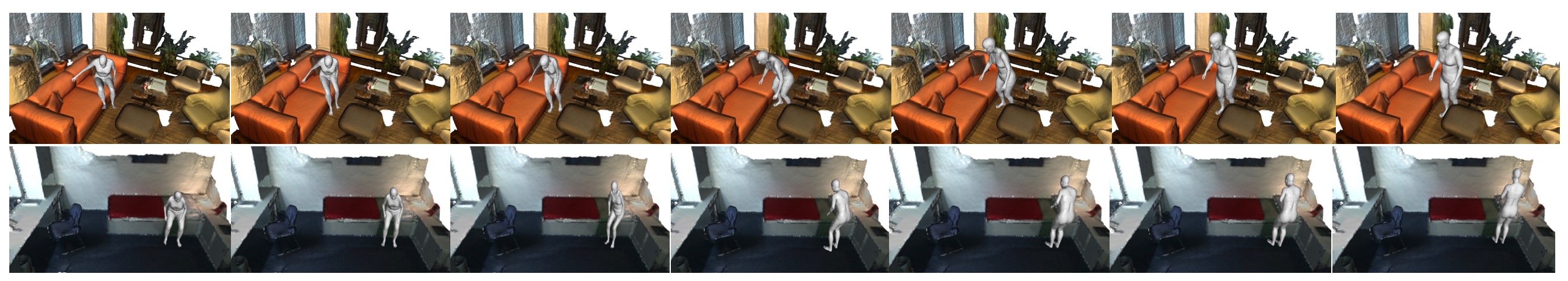}
    \caption{Latent space interpolation. In each row, the left and right images denote two different random samples, 
    and middle images correspond to interpolated latent variables. For each row, the scene BPS $\mathcal{V}_s$ and the scene feature ${\bm x}_s$ are fixed. }
    \label{fig:interpolation}
\end{figure*}

Fig.~\ref{fig:generated_sample} shows some generated human bodies in the three datasets.
To further analyze our cVAE, we perform interpolation in its latent space, and show the results in Fig.~\ref{fig:interpolation}. It can be seen that the body changes smoothly, while preserving natural interactions with the scene. Note that the results in Fig.~\ref{fig:interpolation} are directly produced by the body vertex regressor $H$, without any optimization-based refinement.

\subsection{Limitation and Failure Cases}
Despite the effectiveness verified by our evaluations, our method still has limitations. 
For example, the generated body might inter-penetrate with thin structures in the scene mesh, e.g.~table surface. In this case, \mysf{AdvOptim} cannot resolve this issue, since the value of the collision loss $\mathcal{L}_\textrm{coll}$ is already very small. We thus expect a novel collision representation in future could resolve this issue. Furthermore, detailed and diverse hand poses can not be well synthesized by the current model. Because the basis point set on the scene mesh are too sparse to model the hand-object interaction. We expect a coarse-to-fine encoding scheme could help. We consider these challenges as exciting future directions. Furthermore, the visualizations of the failure cases can be found in the Appendix (Fig.~\ref{fig:supp-failure-case}).

\subsection{Ablation Study}
\label{sec:ablation}

We investigate the influence of the training loss and the optimization scheme on our model performance on {\bf PROX} dataset. More ablation studies are referred to the Appendix.

\begin{table}[t!]
\centering
\footnotesize
\caption{Ablation study w.r.t. diversity metrics.}
\begin{tabular}{llcc}
  \toprule[1pt]
  Optimization & Method &  entropy & cluster size \\
  \midrule

  \multirow{4}*{\mysf{SimOptim}} 
  & \mysf{Ours} & 2.89  & 2.56  \\
  & \mysf{Ours} + $\mathcal{L}_\textrm{contact}$ & \textbf{2.91}  & \textbf{2.59} \\
  & \mysf{Ours} + $\mathcal{L}_\textrm{coll}$ & 2.89  & 2.50  \\
  & \mysf{Ours} + $\mathcal{L}_\textrm{contact}$ + $\mathcal{L}_\textrm{coll}$ & 2.90  & 2.48 \\
  \midrule
  
  \multirow{4}*{\mysf{AdvOptim}} 
  & \mysf{Ours} & 2.90 & 2.67 \\
  & \mysf{Ours} + $\mathcal{L}_\textrm{contact}$ & \textbf{2.91} & \textbf{2.72} \\
  & \mysf{Ours} + $\mathcal{L}_\textrm{coll}$ & 2.88 & 2.63 \\
  & \mysf{Ours} + $\mathcal{L}_\textrm{contact}$ + $\mathcal{L}_\textrm{coll}$ & 2.89 & 2.61  \\
  \bottomrule[1pt]
 \end{tabular}
\label{tab:ablation-loss-diversity}
\end{table}

\begin{table}[t!]
\centering
\footnotesize
\caption{Ablation study w.r.t. physical plausibility metrics.}
\begin{tabular}{llcc}
  \toprule[1pt]
  Optimization & Method &  non-coll & contact  \\
  \midrule
  
  \multirow{4}*{w/o} 
  & \mysf{Ours} & \textbf{0.95}  & 0.83 \\
  & \mysf{Ours} + $\mathcal{L}_\textrm{contact}$ & \textbf{0.95} & \textbf{0.90} \\
  & \mysf{Ours} + $\mathcal{L}_\textrm{coll}$ & \textbf{0.95} & 0.83  \\
  & \mysf{Ours} + $\mathcal{L}_\textrm{contact}$ + $\mathcal{L}_\textrm{coll}$ & \textbf{0.95} & 0.88 \\
  \midrule
  
  \multirow{4}*{\mysf{SimOptim}} 
  & \mysf{Ours}  & \textbf{0.96} & 0.84 \\
  & \mysf{Ours} + $\mathcal{L}_\textrm{contact}$  & \textbf{0.96} & \textbf{0.95} \\
  & \mysf{Ours} + $\mathcal{L}_\textrm{coll}$ & \textbf{0.96} & 0.87 \\
  & \mysf{Ours} + $\mathcal{L}_\textrm{contact}$ + $\mathcal{L}_\textrm{coll}$ & 0.95 & 0.94 \\
  \midrule
  
  \multirow{4}*{\mysf{AdvOptim}} 
  & \mysf{Ours}  & \textbf{0.98} & \textbf{0.99}\\
  & \mysf{Ours} + $\mathcal{L}_\textrm{contact}$  & \textbf{0.98} & \textbf{0.99}\\
  & \mysf{Ours} + $\mathcal{L}_\textrm{coll}$  & \textbf{0.98} & 0.98 \\
  & \mysf{Ours} + $\mathcal{L}_\textrm{contact}$ + $\mathcal{L}_\textrm{coll}$ & \textbf{0.98} & \textbf{0.99} \\
  \bottomrule[1pt]
 \end{tabular}
\label{tab:ablation-loss-physical}
\end{table}

Tab.~\ref{tab:ablation-loss-diversity} and Tab.~\ref{tab:ablation-loss-physical} present model performances when trained with or without contact loss $\mathcal{L}_\textrm{contact}$ and collision loss $\mathcal{L}_\textrm{coll}$ in Eq. \eqref{eqn:training loss}, respectively.
We find that the model without $\mathcal{L}_\textrm{contact}$ and $\mathcal{L}_\textrm{coll}$ has inferior contact scores, indicating that it tends to produce more floating bodies. Also, in contrast to Zhang et al. \cite{zhang2020generating}, we find the loss term $\mathcal{L}_\textrm{contact}$ is effective to improve the contact score in our model.  
Additionally, the model trained without $\mathcal{L}_\textrm{coll}$ already delivers high non-collision scores, indicating the effectiveness of our BPS-based to avoid inter-penetration. However, $\mathcal{L}_\textrm{coll}$ in training reduces the cluster size. 
It is possibly because there exists inter-penetration cases in the ground truth data, which causes contradictions between $\mathcal{L}_\textrm{coll}$ and reconstruction loss during training.
Thus, we mostly use \mysf{Ours} trained with $\mathcal{L}_\textrm{contact}$ to compare with baseline methods.
Moreover, when comparing the results between \mysf{SimOptim} and \mysf{AdvOptim}, we can observe that the collision and contact terms in Eq.~\eqref{eqn:optimization loss} significantly improve both physical plausibility and diversity metrics, which is consistent with \cite{zhang2020generating}.

\section{Conclusion}
\label{sec:conclusion}
This paper introduces a novel representation to model human-scene interactions by explicitly encoding the proximity between the body and the environment. It is purely geometry-based and consistent across various bodies and scene structures, thus only requiring the 3D mesh as the input. 
By training a generative model to synthesize such representations and body meshes, our approach is able to generate humans in natural interactions with new environments, which demonstrates the practical merits of our work. 

{\small
\myparagraph{Acknowledgements.}
We sincerely thank Mohamed Hassan for his scene SDF calculation protocol. This work was partially supported by the German Research Foundation (DFG): SFB 1233, Robust Vision: Inference Principles and Neural Mechanisms, TP XX, project number: 276693517. Qianli Ma acknowledges the support from the Max Planck ETH Center for Learning Systems.
}

{\small \myparagraph{Disclosure.}
MJB has received research gift funds from Intel, Nvidia, Adobe, Facebook, and Amazon. While MJB is a part-time employee of Amazon, his research was performed solely at MPI. He is also an investor in Meshcapde GmbH.}

{\small
\bibliographystyle{ieee}
\bibliography{egbib}
}

\begingroup
\onecolumn 

\appendix
\begin{center}
\Large{\bf PLACE: Proximity Learning of Articulation and Contact in 3D Environments \\ **Appendix**}
\end{center}

\setcounter{page}{1}
\setcounter{table}{0}
\setcounter{figure}{0}
\renewcommand{\thetable}{S\arabic{table}}
\renewcommand\thefigure{S\arabic{figure}}

\section{Architecture Details}
The detailed architecture is illustrated in Fig.~\ref{fig:supp-network-detail}.
During the training stage, the scene-conditioned VAE takes the body feature ${\bm x}_b$, and reconstructs the body feature ${\bm x}^{rec}_b$ based on the scene feature ${\bm x}_s$. Residual blocks are utilized in the scene-conditioned VAE, and the middle latent feature of the scene feature auto-encoder $E$ is concatenated into different layers in $G$ to enforce stronger scene conditioning. During inference, the scene-conditioned VAE generates a body feature from the randomly sampled noise $z$ conditioned on the scene feature input ${\bm x}_s$.

The body mesh vertex regressor includes an auto-encoder $F$ for the scene BPS $\mathcal{V}_s$, and a multilayer perceptron $H$ to reconstruct the vertex coordinates of the final body mesh. ${\Delta \mathcal{V}}^{rec}_b$ denotes a global 3D translation to improve contact relationship.

\begin{figure}[h!]
    \centering
    \includegraphics[width=15cm]{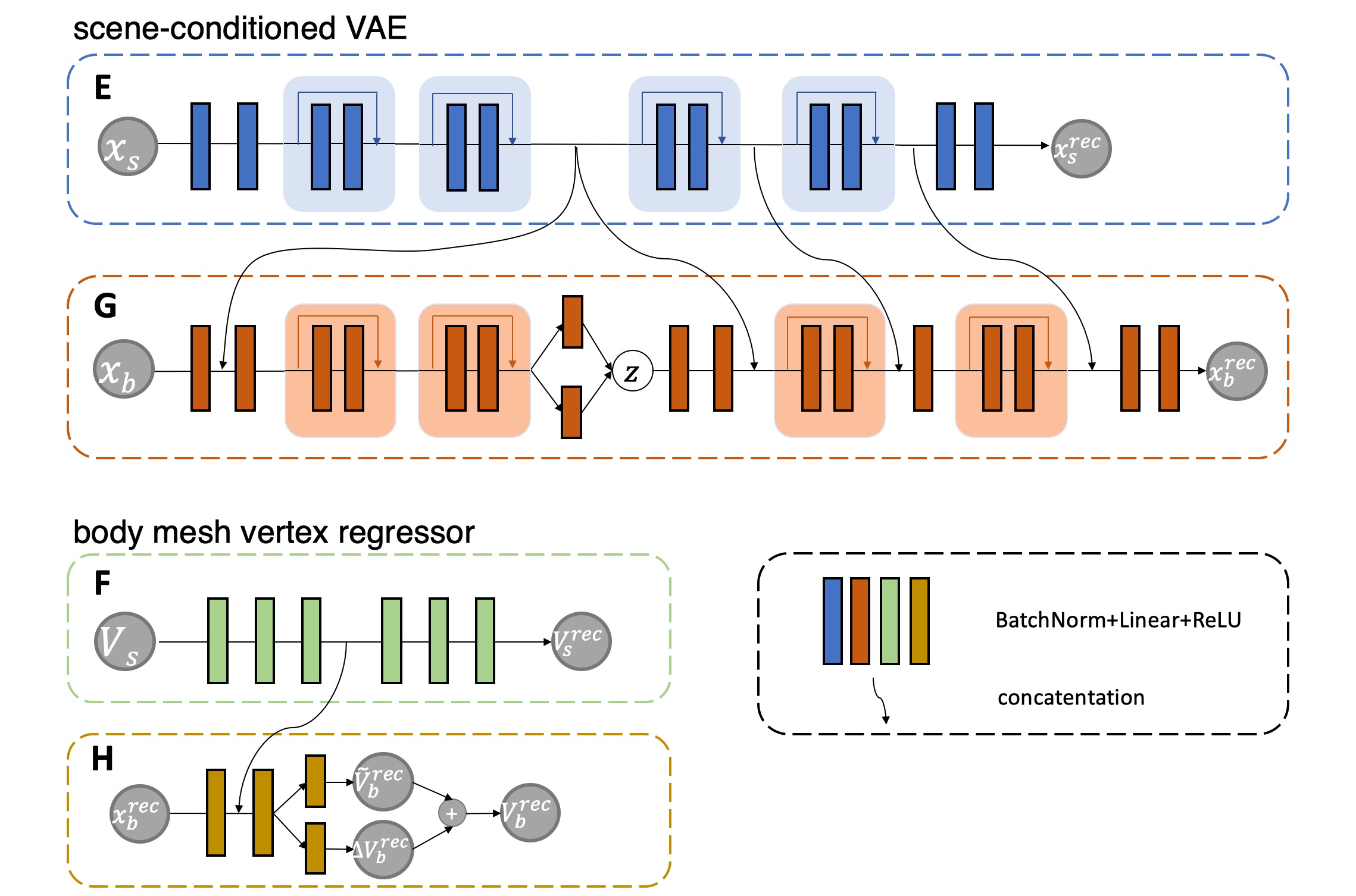}
    \caption{Architecture details. ${\bm x}_s$, $\mathcal{V}_s$, ${\bm x}_b$ and $\mathcal{V}_b$ denote the scene feature, the scene BPS, the body feature, and the full body mesh vertices, respectively.  ${\Delta \mathcal{V}}^{rec}_b$ denotes a global 3D translation.}
    \label{fig:supp-network-detail}
\end{figure}

\section{Experiment Details}
\subsection{Implementation Details} 
We use PyTorch \cite{paszke2017automatic} for implementation. We use ADAM \cite{KingmaB14} as the optimizer ($\beta_1=0.9$, $\beta_2=0.999$), with the learning rate 1e-4. The 3D cage size for BPS encoding (see Sec.~\ref{sec:bps}) is set to 2m$^3$ by default.
We use the same Chamfer distance implementation in the contact loss as in \cite{deprelle2019learning, groueix2018b, PROX:2019, zhang2020generating}. 
The training loss weights in Eq.~\ref{eqn:training loss} are set as $\alpha_1 = 0.5$, with $\{\alpha_2, \alpha_3\}$ set to $\{0.001, 0.001\}$ and enabled after 75\% of the training epochs.
The loss weights for optimization in Eq.~\ref{eqn:optimization loss} are empirically set as $\lambda_1 = 8$ (PROX, MP3D), $\lambda_1 = 0.1$ (Replica), $\lambda_2 = 0.5$, $\lambda_3 = 0.02$, $\lambda_4 = 0.01$, $\lambda_5 = 0.01$.
The collision loss weight $\lambda_1$ is set to a much smaller weight for Replica as the corresponding scenes are larger than PROX and MP3D, with a higher ratio of generated humans in standing poses, which seldom suffers from inter-penetration. 
We train the model for around 260 epochs on a single GeForceGTX1080Ti GPU, with a batch size of 120.

\subsection{Ablation Study}
\label{append_ablation_study}

Here we investigate model performance from the perspectives of: (1) model architecture, (2) regressing SMPL-X parameters, and (3) 3D cage size selection.

\begin{table*}[t!]
\centering
\caption{\textbf{Ablation study for architecture:}
we evaluate the diversity and physical metrics on the PROX dataset with different architecture variants, 
where \textbf{w/o ML-cat} denotes the model without multi-layer additional intermediate feature concatenation between the decoders of $E$ and $G$ (see Fig.~\ref{fig:network}), 
\textbf{w/o $\bm{E/F}$-dec} denotes the model without decoders in $E$ and $F$,
\textbf{w/o $\bm{E}$} denotes the model without $E$, which takes the latent vector from $F$ into $G$ as scene conditions,
\textbf{$\bm{H}$-S} regresses 72-d SMPL-X parameters instead of body vertices in $H$. 
All models are trained without $\mathcal{L}_{contact}$ and collision loss $\mathcal{L}_{coll}$ in Eq.~\eqref{eqn:training loss}.
}
\begin{tabular}{lccccc}
  \toprule[1pt]
  & & \multicolumn{2}{c}{Physical plausibility} & \multicolumn{2}{c}{Diversity} \\
  \cmidrule(lr){3-4} \cmidrule(lr){5-6}
  Optimization & Model &  non-coll & contact & entropy & cluster size  \\
  \midrule[1pt]
  \multirow{5}*{w/o} 
  & w/o ML-cat & \textbf{0.95} & 0.78 & - & -  \\
  & w/o $E$/$F$-dec & \textbf{0.95} & 0.82 & - & - \\
  & w/o $E$ & \textbf{0.95} & 0.79 & - & - \\
  & $H$-S & \textbf{0.95} & \textbf{0.92} & 2.94 & 1.68 \\
  & \mysf{Ours} & \textbf{0.95} & 0.83 & - & - \\
  \midrule

  \multirow{5}*{\mysf{SimOptim}} 
  & w/o ML-cat  & \textbf{0.96} & \textbf{0.84} & \textbf{2.89} & 2.41 \\
  & w/o $E$/$F$-dec  & \textbf{0.96} & \textbf{0.84} & 2.85 & 2.42 \\
  & w/o $E$  & 0.95 & \textbf{0.84} & \textbf{2.89} & 2.54 \\
  & \mysf{Ours}  & \textbf{0.96} & \textbf{0.84} & \textbf{2.89} & \textbf{2.56} \\
  \midrule 
  
  \multirow{5}*{\mysf{AdvOptim}} 
  & w/o ML-cat  & \textbf{0.98} & 0.97 & 2.87 & 2.52 \\
  & w/o $E$/$F$-dec  & \textbf{0.98} & 0.98 & 2.87 & 2.52 \\
  & w/o $E$  & 0.97 & 0.98 & 2.88 & 2.64 \\
  & $H$-S  & 0.97 & \textbf{0.99} & \textbf{2.95} & 1.99 \\
  & \mysf{Ours} & \textbf{0.98} & \textbf{0.99} & 2.90 & \textbf{2.67} \\
  \toprule[1pt]
 \end{tabular}
\label{tab:supp-ablation}
\end{table*}

\paragraph{Model architecture. }
Tab.~\ref{tab:supp-ablation} studies the following components of the proposed architecture: the additional intermediate feature concatenation between the decoders of $E$ and $G$ (see Fig.~ \ref{fig:network}), the decoders in $E$ and $F$, and the scene feature auto-encoder $E$. The scene BPS auto-encoder $F$ cannot be removed as the model needs coordinate-system-dependent information to regress body vertex locations.
The cluster size decreases in the absence of the first two components (w/o ML-cat and w/o $E/F$-dec), as these two components aim to enforce stronger scene conditioning for reconstruction and generation, hence are significant for diversity.
The model without $E$ achieves comparable scores with the full model, but we propose to encode the scene by BPS representation instead of coordinates in c-VAE due to its simplicity and independence of the coordinate system, yielding better generalization abilities.

\paragraph{Regressing SMPL-X parameters. }
As shown in Tab.~\ref{tab:supp-ablation}, although the model regressing SMPL-X parameters ($H$-S) instead of body vertex coordinates achieve good contact scores, it suffers severely degraded performance w.r.t. cluster size, indicating that the generated bodies are more centralized within each cluster. Since the body mesh regressor $H$ is conditioned on coordinates of scene BPS, regressing body vertex coordinates is an easier task than regressing SMPL-X parameters, and can learn more diverse details of body poses.

\paragraph{3D cage size selection. }
See results of models trained with different cage sizes in Tab.~\ref{tab:supp-cage-size}.
One obvious pattern can be observed is that the cluster size in diversity metrics increases with the cage size decreasing. This is probably because smaller cage sizes increase the randomness in global positions of the generated bodies, as the cage is randomly cropped on test scenes.
We choose the size for the 3D cage following 3 principles: 1) the size should be large enough to contain the full body in different poses, 2) the size should only include scene objects near the body, 3) the size should not be larger than training scenes, therefore we set the 3D cage size as 2m$^3$.

\begin{table}
\centering
\caption{\textbf{Ablation study for 3D cage size:} the performance metrics for the model \mysf{Ours} + $\mathcal{L}_\textrm{contact}$ trained with different cage sizes.}
\begin{tabular}{lccccc}
  \toprule[1pt]
  & & \multicolumn{2}{c}{Physical plausibility} & \multicolumn{2}{c}{Diversity} \\
  \cmidrule(lr){3-4} \cmidrule(lr){5-6}
  Optimization & Cage size &  non-coll & contact & entropy & cluster size  \\
  \midrule[1pt]
  
  \multirow{4}*{w/o} 
  & 1.5 & 0.92 & 0.90  & - & - \\
  & 2.0 & \textbf{0.95} & 0.90  & -  & - \\
  & 2.5 & 0.93 & \textbf{0.95}  & -  & - \\
  & 3.0 & 0.92 & \textbf{0.95}  & -  & - \\
  \midrule
  
  \multirow{4}*{\mysf{SimOptim}} 
  & 1.5 & 0.93 & 0.95 & \textbf{2.93} & \textbf{2.93} \\
  & 2.0 & \textbf{0.96} & 0.95 & 2.91 & 2.59 \\
  & 2.5 & 0.93 & 0.95 & 2.92 & 2.38 \\
  & 3.0 & 0.92 & \textbf{0.96} & 2.91 & 2.36 \\
  \midrule
  
  \multirow{4}*{\mysf{AdvOptim}} 
  & 1.5 & 0.96 & 0.98 & \textbf{2.91} & \textbf{3.09} \\
  & 2.0 & \textbf{0.98} & \textbf{0.99} & \textbf{2.91} & 2.72\\
  & 2.5 & 0.96 & \textbf{0.99} & 2.88 & 2.55\\
  & 3.0 & 0.95 & \textbf{0.99} & 2.90 & 2.53\\
  \bottomrule[1pt]
 \end{tabular}
\label{tab:supp-cage-size}
\end{table}

\subsection{Data Augmentation}
The data preprocessing involves two different coordinate systems, i.e. the 3D cage coordinate and the unit sphere coordinate.
Given a (scene, body) pair of meshes from training data, the (scene, body) pair is randomly rotated around $z$ axis of the original world coordinate. A 3D cage of 2m$^3$ is randomly selected around the body, and the cage center is defined as following: given a body mesh, in $x, y$ axis, the body center is selected as the initial cage center, then we randomly shift the cage center by $\Delta{x} \in [-\frac{r}{3}, \frac{r}{3}]$, $\Delta{y} \in [-\frac{r}{3}, \frac{r}{3}]$ in $x, y$ axis respectively. In $z$ axis, the cage center is fixed at $1$m above the ground. 
Then the scene mesh is trimmed accordingly. The (scene cage, body) vertices are rescaled by the cage size into the unit sphere, with the cage center positioned at the unit sphere center.
The \textbf{3D cage coordinate} is then defined with the origin placed at the rescaled cage center.
The \textbf{unit sphere coordinate} is introduced to derive the distance-based BPS features (i.e. ${\bm x}_s$ and ${\bm x}_b$), with the unit sphere center as the origin. For each training sample, the (scene cage, body) is randomly rotated around $z$ axis of the unit sphere, and randomly shifted in a small range within the unit sphere to provide more diversity for training. Note that the point coordinates $\mathcal{V}_s$ and $\mathcal{V}_b$ are relative to the 3D scene cage coordinate.

With such data augmentation, the relative position between the body and 3D scene cage, and between the 3D scene cage and unit sphere will be different for each sample. This pipeline is repeated 4 times for each (scene, body) pair to enlarge the size of the training set.
During inference, a 3D scene cage of the same size is randomly trimmed from the test scene.

\subsection{Details of the User Studies}
In order to evaluate the perceptual naturalness of the interactions between the generated humans and the environment, we run user study on Amazon Mechanical Turk (AMT) with two protocols: (1) unary user study, and (2) binary user study.
For each scene-human pair, we display two images rendered from two different camera views to the users. 

In the unary user study, each user is presented with one generation result each time, and asked to give a score between 1 (strongly not natural) and 5 (strongly natural). To alleviate randomness in the users' ratings, each result is rated by 3 users in the unary study, and we re-evaluate the baseline methods.

In addition, we run the binary user study to compare between our method and the baseline method in \cite{zhang2020generating}, as well as the ground truth of PROX dataset. The users are presented with two results each time, one from the proposed method, and the other one from the baseline method or the ground truth. The user will select one which they think is more natural. The user interfaces for both unary and binary study are shown in Fig.~\ref{fig:supp-user-study}.

\begin{figure}[h!]
    \centering
    \includegraphics[width=10cm]{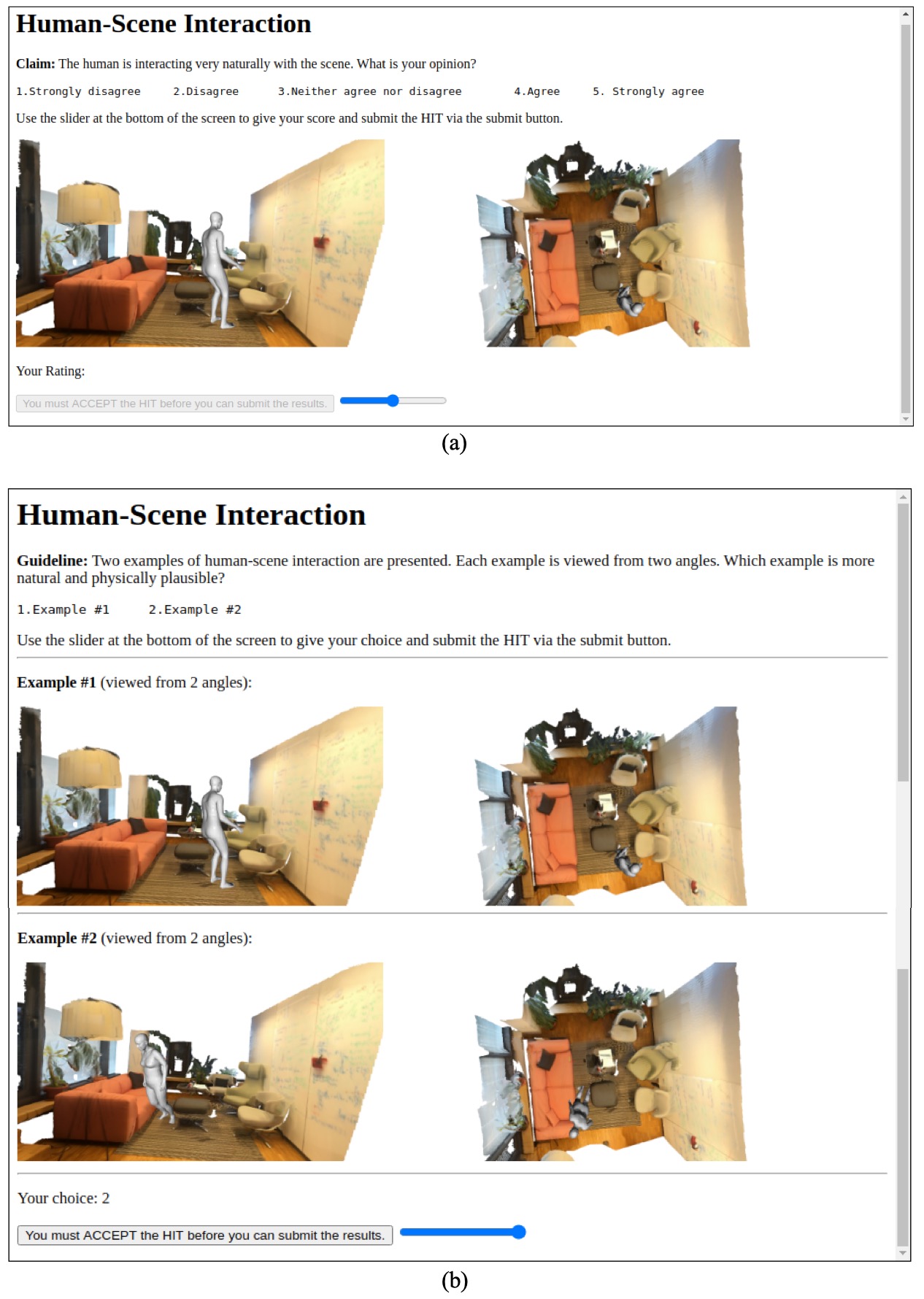}
    \caption{The user interface of the user study. (a) Unary study: the user is requested to give a score between 1 and 5 to evaluate how natural the huamn-scene interaction is. (b) Binary study: the user is requested to select the method in which the result is more natural.}
    \label{fig:supp-user-study}
\end{figure}

\subsection{Examples of the SDFs for MP3D}
We find that the SDFs on MP3D are not sufficiently reliable, which is probably due to noisy scans. Fig.~\ref{fig:supp-mp3d-sdf-example} shows an example where the contact score is miscalculated as 0 by the corresponding SDF, suggesting that there is no human-scene contact, even with the obvious contact between the human and the scene (Fig.~\ref{fig:supp-mp3d-sdf-example} (b)). 

\begin{figure}[h!]
    \centering
    \includegraphics[width=10cm]{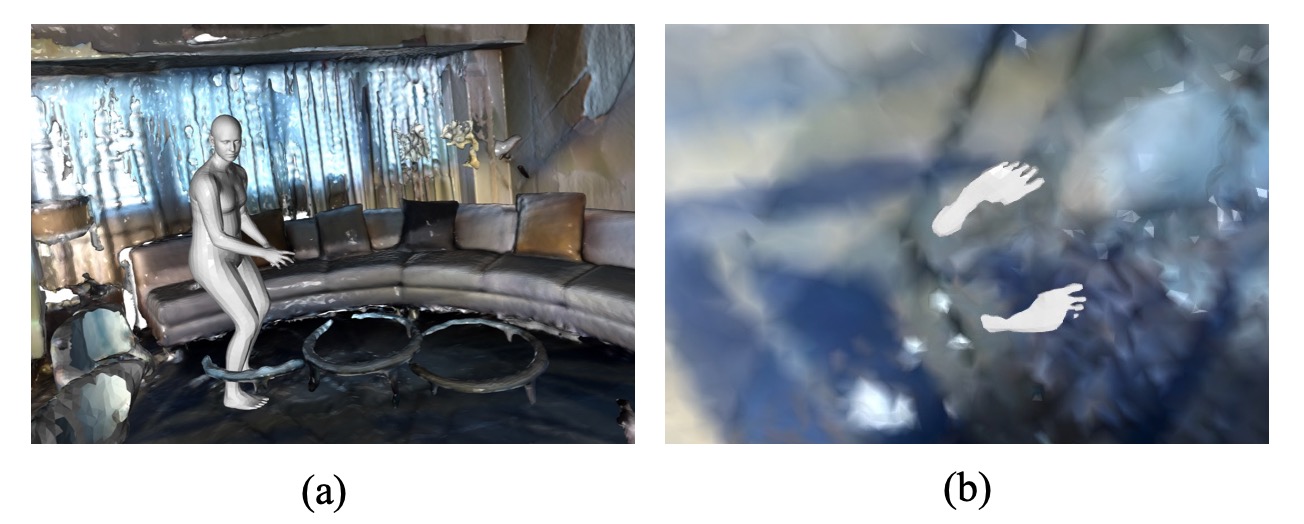}
    \caption{An example of noisy ground truth SDF from the MP3D dataset that leads to inaccurate contact metric evaluation: (a) a generated body mesh in the scene `17DRP5sb8fy-livingroom' of MP3D dataset, (b) bottom-up view of the scene and body: the feet clearly have contact with the ground, however the the contact score is miscalculated as 0 using the ground truth SDF.}
    \label{fig:supp-mp3d-sdf-example}
\end{figure}

\section{More Qualitative Results and Failure Cases}
Fig.~\ref{fig:supp-quality-prox}, Fig.~\ref{fig:supp-quality-replica} and Fig.~\ref{fig:supp-quality-mp3d} show qualitative results of the proposed method on the test scenes of PROX, Replica and MP3D datasets respectively.

\begin{figure}[h!]
    \centering
    \includegraphics[width=15cm]{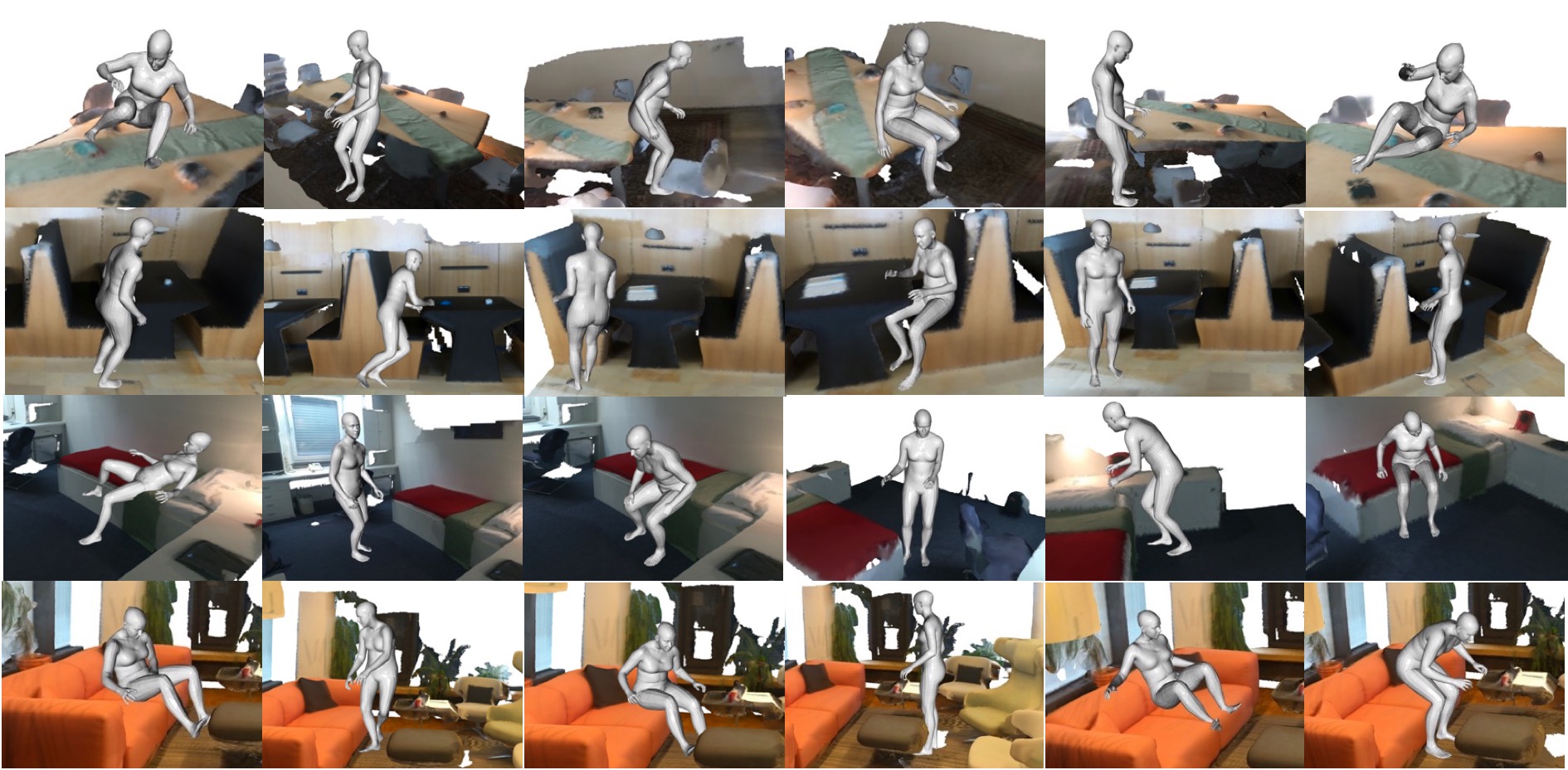}
    \caption{Qualitative results on test scenes of PROX dataset.}
    \label{fig:supp-quality-prox}
\end{figure}

\begin{figure}[h!]
    \centering
    \includegraphics[width=15cm]{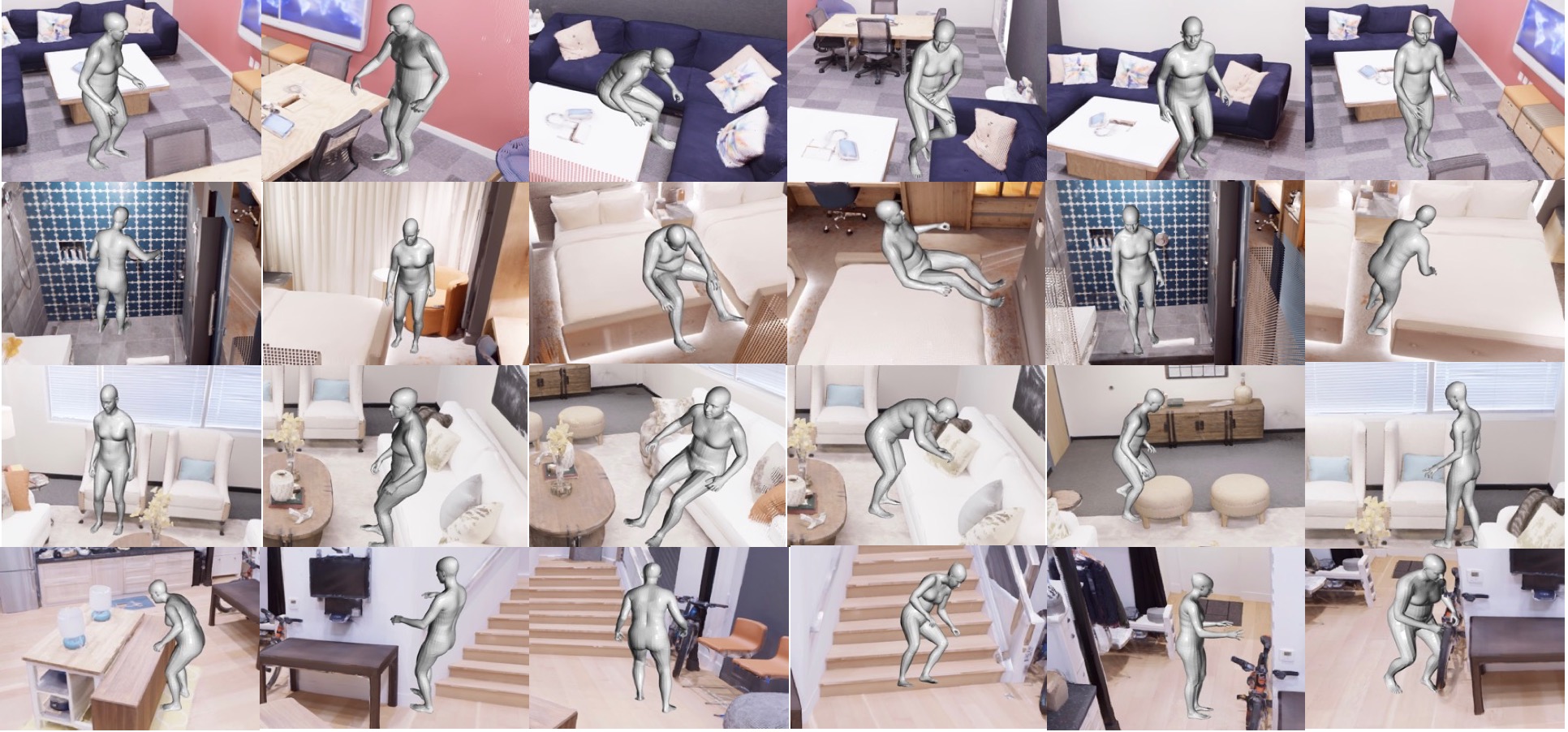}
    \caption{Qualitative results on Replica dataset.}
    \label{fig:supp-quality-replica}
\end{figure}

\begin{figure}[h!]
    \centering
    \includegraphics[width=15cm]{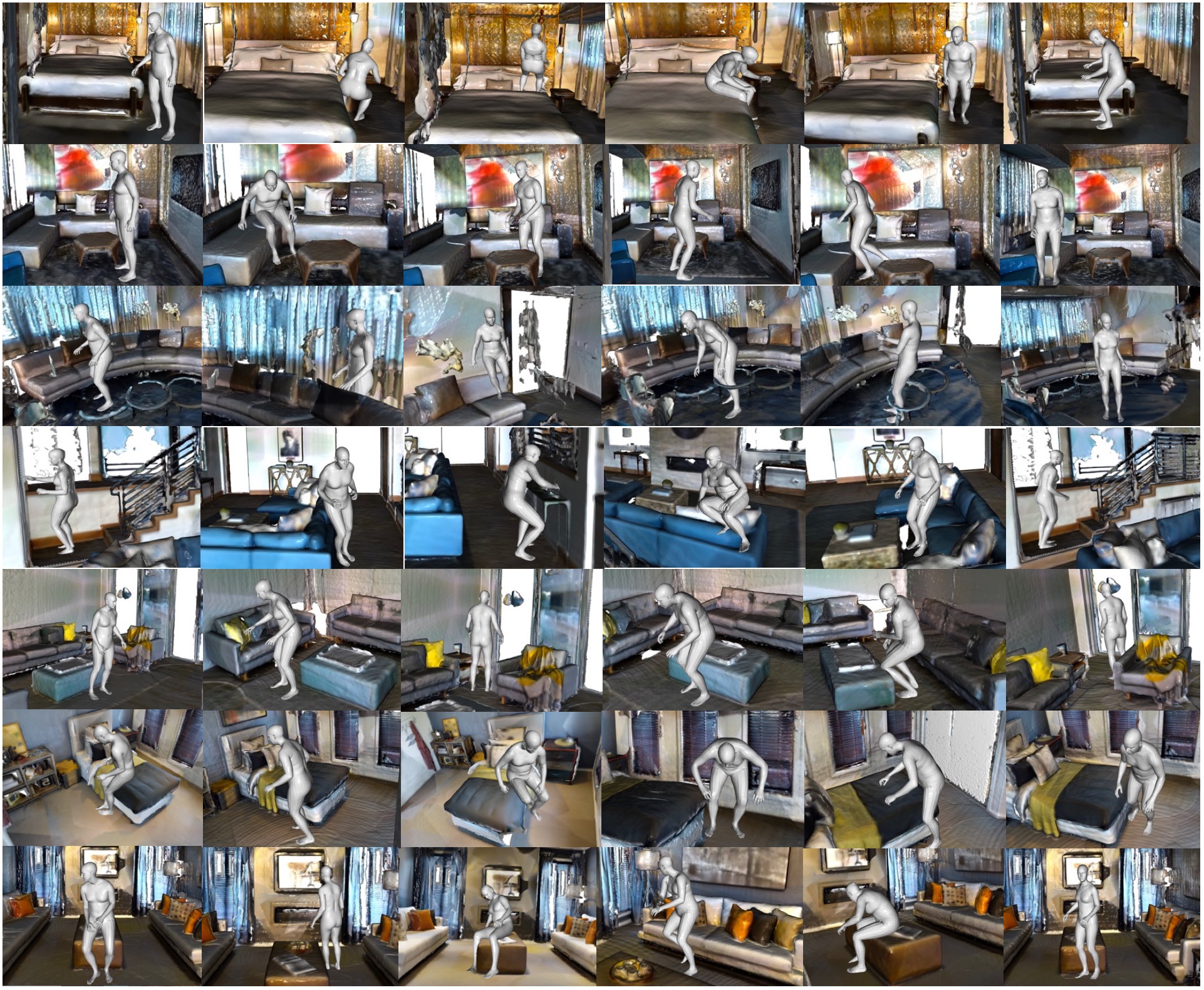}
    \caption{Qualitative results on MP3D dataset.}
    \label{fig:supp-quality-mp3d}
\end{figure}

The failure cases are illustrated in Fig.~\ref{fig:supp-failure-case}, which can be classified into three cases. 
(1) The inter-penetration between the generated body before optimization is severe. Interaction-based optimization can hardly solve such case (Fig.~\ref{fig:supp-failure-case} (a)) since it is non-trivial to deal with this inter-penetration case while reconstructing the generated body vertices and body features.
(2) The generated body before optimization inter-penetrates with a thin structure in the scene, such as table or chair. The corresponding collision loss value in the interaction-based optimization is quite small, therefore this case can hardly be improved by the optimization (Fig.~\ref{fig:supp-failure-case} (b)).
(3) The generated body is floating with a small distance above the ground, or with slight inter-penetration with the scene mesh before optimization. There are two possible reasons for the floating. Firstly, there are floating bodies in the ground truth of PROX dataset, which can cause negative influences in training. Secondly, the generated body feature of the body parts in contact with the scene may not be exactly zero. The interaction-based optimization can improve the performance in this case by the contact term and collision term respectively ((Fig.~\ref{fig:supp-failure-case} (c))).

\begin{figure}[h!]
    \centering
    \includegraphics[width=\linewidth]{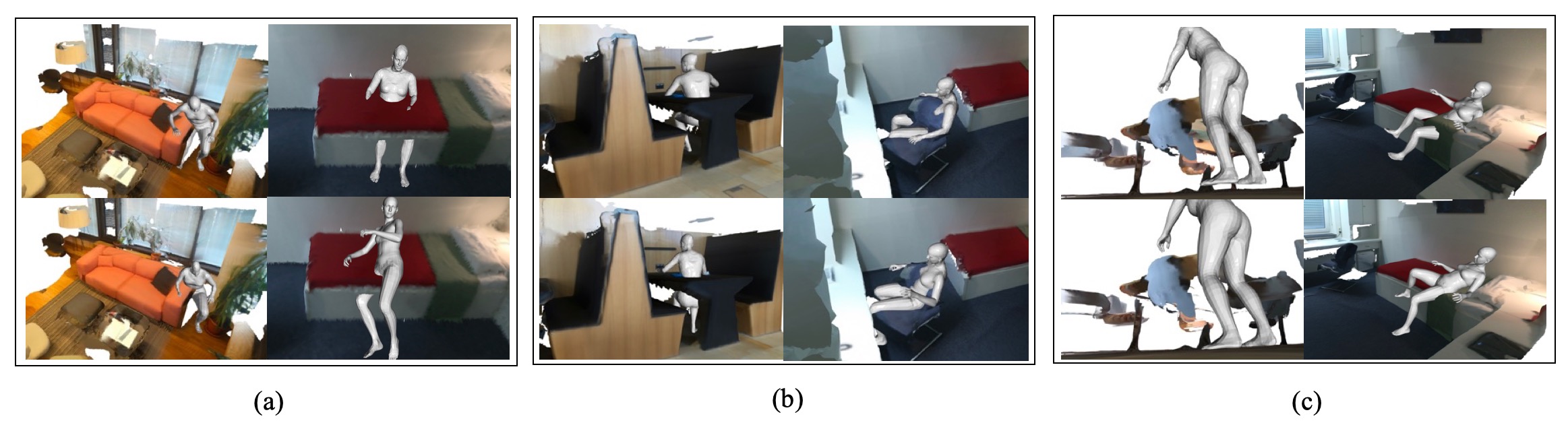}
    \caption{Failure cases: (a) severe inter-penetrations with the scene mesh, (b) inter-penetrations with thin structures in the scene and, (c) floating bodies or slight inter-penetrations with the scene mesh. The first row denotes results without interaction-based optimization, and the second row denotes results after interaction-based optimization.}
    \label{fig:supp-failure-case}
\end{figure}

\end{document}